\documentclass{article}

\usepackage[accepted]{icml2021}
\usepackage{microtype}
\usepackage{subfigure}
\usepackage{booktabs} %

\usepackage{algorithm}
\usepackage{algorithmic}
\usepackage[inline]{enumitem}
\usepackage{amsfonts}  %
\usepackage{courier}
\usepackage{amsmath}
\usepackage{graphbox}
\usepackage{graphicx}
\usepackage{placeins}
\usepackage{fontawesome}

\usepackage{hyperref}
\usepackage{comment}
\usepackage{comment}
\usepackage{csquotes}

\icmltitlerunning{Neural Symbolic Regression that Scales}

\begin{document}

\twocolumn[
\icmltitle{Neural Symbolic Regression that Scales}

\icmlsetsymbol{equal}{*}

\begin{icmlauthorlist}
\icmlauthor{Luca Biggio}{equal,to,goo}
\icmlauthor{Tommaso Bendinelli}{equal,goo}
\icmlauthor{Alexander Neitz}{ed}
\icmlauthor{Aurelien Lucchi}{to}
\icmlauthor{Giambattista Parascandolo}{to,ed}
\end{icmlauthorlist}

\icmlaffiliation{to}{Department of Computer Science, ETH, Zürich, Switzerland}
\icmlaffiliation{goo}{CSEM SA, Alpnach, Switzerland}
\icmlaffiliation{ed}{Max Planck Institute for Intelligent Systems, Tübingen, Germany}
\icmlcorrespondingauthor{Luca Biggio}{luca.biggio@inf.ethz.ch}
\icmlcorrespondingauthor{Tommaso Bendinelli}{tommaso.bendinelli@csem.ch}

\icmlkeywords{Machine Learning, ICML}

\vskip 0.3in
]

\printAffiliationsAndNotice{\icmlEqualContribution} %

\begin{abstract}
Symbolic equations are at the core of scientific discovery.
The task of discovering the underlying equation from a set of input-output pairs is called symbolic regression.
Traditionally, symbolic regression methods use hand-designed strategies that do not improve with experience.
In this paper, we introduce the first symbolic regression method that leverages large scale pre-training.
We procedurally generate an unbounded set of equations, and simultaneously pre-train a Transformer to predict the symbolic equation from a corresponding set of input-output-pairs.
At test time, we query the model on a new set of points and use its output to guide the search for the equation.
We show empirically that this approach can re-discover a set of well-known physical equations, and that it improves over time with more data and compute.

\end{abstract}

\section{Introduction}
\vspace{-0.5mm}

Since the early ages of Natural Sciences in the sixteenth century, the process of scientific discovery has rooted in the formalization of novel insights and intuitions about the natural world into compact symbolic representations of such new acquired knowledge, namely, mathematical equations.
\vspace{-0.5mm}

Mathematical equations encode both objective descriptions of experimental data and our inductive biases about the regularity we attribute to natural phenomena.
When seen under the perspective of modern machine learning, they present a number of appealing properties: 
\begin{enumerate*}[label=(\roman*)]
\item They provide \textit{compressed} and \textit{explainable} representations of complex phenomena.
\item They allow to easily incorporate prior knowledge.
\item When relevant aspects about the data generating process are captured, they often generalize well beyond the distribution of the observations from which they were derived.
\end{enumerate*}

The process of discovering symbolic expressions from experimental data is hard and has traditionally been one of the hallmarks of human intelligence. 
\textit{Symbolic regression} is a branch of regression analysis that tries to emulate such a process. 
More formally, given a set of $n$ input-output pairs $\{(x_i, y_i)\}_{i=1}^n\sim\mathcal{X}\times\mathcal{Y}$, the goal is to find a symbolic equation $e$ and corresponding function $f_e$ such that $y \approx f_e(x)$  for all ${(x,y)}\in\mathcal{X}\times\mathcal{Y}$. In other words, the goal of symbolic regression is to infer both model structure and model parameters in a data-driven fashion. Even assuming that the vocabulary of primitives --- e.g. $\{\sin, \exp, +$, ...\} --- is sufficient to express the \textit{correct} equation behind the observed data, symbolic regression is a hard problem to tackle.
The number of functions associated with a string of symbols grows exponentially with the string length, and the presence of numeric constants further exacerbates its difficulty.
\vspace{-0.2mm}

Due to its challenging combinatorial nature, existing approaches to symbolic regression are mainly based on search-techniques whose goal is typically to minimize a pre-specified fitness function measuring the distance between the predicted expression and the available data.
The two main drawbacks of such methods are that:
\begin{enumerate*}[label=(\roman*)]
\item \textit{They do not improve with experience.} 
As every equation is regressed from scratch, the system does not improve if access to more data from different equations is given.
\item \textit{The inductive bias is opaque.} It is difficult for the user to steer the prior towards a specific class of equations (e.g. polynomials, etc.). In other words, even though most symbolic regression algorithms generate their prediction starting from a fixed set of primitives reflecting the user's prior knowledge, such elementary building blocks can be combined in many arbitrary ways, providing little control over the equation distribution.
\end{enumerate*}
To overcome both drawbacks, in this paper we take a step back, and let the model \textit{learn the task} of symbolic regression over time, on a user-defined prior over equations. 
\vspace{-0.2mm}

Building on the recent successes of large models trained on large datasets \cite{gpt3,bert,simclr,simclr2}, we show that a strong symbolic regressor can be purely learned from data. 
The key factor behind our approach is that computers can generate unbounded amounts of data with perfect accuracy and at virtually no cost. The distribution over equations used during pre-training strongly influences the prior over equations of the final system. 
Such a prior thus becomes easy to understand and control.

The main contributions of this paper are the following:
\vspace{-3mm}
\begin{itemize}
    \item We introduce a simple, flexible, and powerful framework for symbolic regression, the first approach (to the best of our knowledge) to improve over time with data and compute.
    \vspace{-0.25mm}
    \item We demonstrate that \textit{learning the task} of symbolic regression from data is sufficient to significantly outperform state-of-the-art approaches relying on hand-designed strategies.
    \vspace{-0.25mm}
    \item {We release our code and largest pre-trained model \footnote{\href{https://github.com/SymposiumOrganization/NeuralSymbolicRegressionThatScales}{https://github.com/SymposiumOrganization/\\NeuralSymbolicRegressionThatScales}}}
\end{itemize}

\vspace{-3mm}
In Section \ref{sec:related_work}, we detail related work in the literature. In Section \ref{sec:method}, we present our algorithm for neural symbolic regression that scales.
We evaluate the method in the experiments described in Section \ref{sec:experiments} and \ref{sec:results} and compare it to state-of-the-art baselines.
In Section \ref{sec:conclusion} we discuss results, limitations, and potential for future work.

\vspace{-2mm}
\section{Related Work}\label{sec:related_work}
\vspace{-0.5mm}

\paragraph{Genetic Programming for Symbolic Regression}
Traditional approaches to symbolic regression are based on genetic algorithms \cite{ga} and, in particular, genetic programming (GP) \cite{gp2}.
GP methods used for symbolic regression iteratively \enquote{evolve} a population of candidate mathematical expressions via mutation and recombination. %
The most popular GP-based technique applied to symbolic regression is undoubtedly the commercial software Eureqa \cite{eureka} which is based on the approach proposed by \citet{gp1}. Despite having shown for the first time the potential of data-driven approaches to the problem of function discovery, GP-based techniques do not scale well to high dimensional problems and are highly sensitive to hyperparameters \cite{brenden}.
\paragraph{Neural Networks for Symbolic Regression}
A more recent line of research explores the potential of deep neural networks to tackle the combinatorial challenge of symbolic regression. \citet{martius2016extrapolation} propose a simple fully-connected neural network where standard activation functions are replaced with symbolic building blocks (e.g. \enquote{$\sin(\cdot)$}, \enquote{$\cos(\cdot)$}, \enquote{$+$}, \enquote{Identity$(\cdot)$}). Once the model is trained, a symbolic formula can be automatically read off from the network architecture and weights. This method inherits the ability of neural networks to deal with high-dimensional data and scales well with the number of input-output pairs. However, it requires specific extensions \cite{SahooLampertMartius2018:EQLDiv} to deal with functions involving divisions between elementary building blocks (e.g. $\frac{sin(x)}{x^2}$) and the inclusion of exponential and logarithmic activations result in exploding gradients and numerical issues.
\vspace{-1mm}

Another approach to circumvent the discrete combinatorial search inherent in the symbolic regression framework is proposed in \cite{GVAE}. Here, a variational autoencoder \cite{VAE} is first trained to reconstruct symbolic expressions and the search for the best fitting function is then performed over the latent space in a subsequent step. While the idea of moving the search for the best expression from a discrete space to a continuous one is interesting and has been exploited by other approaches (e.g. \cite{vds}), the method does not prove to be effective in recovering relatively simple symbolic formulas.
More recently, \citet{brenden} developed a new technique where a recurrent neural network (RNN) is used to model a probability distribution over the space of mathematical expressions. Output expressions contain symbolic placeholders to indicate the presence of numerical constants. Such constants are then fit in a second stage by an out-of-the-box nonlinear optimizer.
The RNN is trained by minimizing a risk-seeking RL objective that assigns a larger reward to the top-epsilon samples from the output distribution. 
The method represents a significant step forward in the application of deep learning to symbolic regression. 
While showing promising results, the network has to be retrained from scratch for each new equation and the RNN is never directly conditioned on the data it is trained to model. %

Finally, neural networks can also be used in combination with existing techniques or hand-designed rules to perform symbolic regression. Notable examples are \cite{AIf1,AIf2}, where neural networks are employed to identify simplifying properties in the data such as additive separability and compositionality. 
These properties are exploited to recursively simplify the original dataset into less challenging sub-problems that can be tackled by a symbolic regression technique of choice. A similar rationale is followed in \cite{Cranmer}, where different components of a trained Graph Neural Network (GNN) are independently fit by a symbolic regression algorithm. By joining the so-found expressions, a final algebraic formula describing the network can be obtained.
The aforementioned approaches might provide very good performances when it is known a priori whether the data are characterized by specific structural properties, such as symmetries or invariances.
However, when such information is not accessible, more domain-agnostic methods are required.\\
\vspace{-0.5cm}

\paragraph{Large Scale Pre-training}
Our approach builds upon a large body of work emphasizing the benefits of pre-training large models on large datasets \cite{scaling,bert,gpt3,simclr,simclr2,overp1}. Examples of such models can be found in Computer Vision \cite{radford2021learning,simclr,simclr2,kolesnikov2020big, CPC} and Natural Language Processing \cite{bert,gpt3}. %
There have also been recent applications of Transformers \cite{transformer} to tasks involving symbolic mathematics manipulations \cite{facebook, mathnn1} and automated theorem proving \cite{polu2020generative}.
Our work builds on the results from \citet{facebook}, where Transformers are trained to successfully perform challenging mathematical tasks such as symbolic integration and solving differential equations.
However, our setting presents the additional challenge of mapping \textit{numerical values} to the corresponding symbolic formula, instead of working exclusively within the symbolic domain.

\vspace{-0.3mm}
\section{Neural Symbolic Regression that Scales}\label{sec:method}
\vspace{-0.25mm}

\begin{figure*}
\centering
  \includegraphics[scale = 0.55]{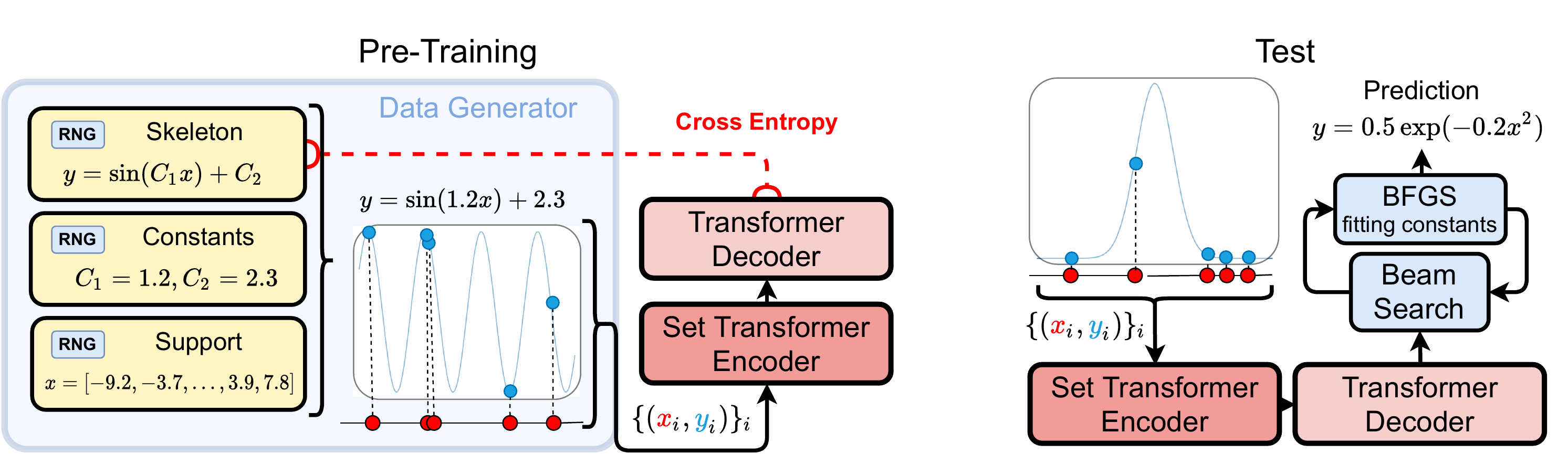}
  \caption{(Left) The data generator produces the input for the Transformer and its target expression. It does so by randomly sampling (i) an equation skeleton (including placeholders for the constants), (ii) numerical constants used to replace the placeholders and (iii) a set of support points $\{x_i\}_i$ to evaluate the previously generated equation and get the corresponding $\{y_i\}_i$. The $\{(x_i, y_i)\}_i$ pairs are fed into the Transformer, which is trained to minimize the cross-entropy loss with the ground-truth skeleton without numerical constants. Both the model output and the targets are expressed in prefix notation. (Right) At test time, given new input data, we sample candidate symbolic skeletons from the model using beam-search. The final candidate equations are obtained by fitting the constants with BFGS.}\label{procedure}
\end{figure*}

A symbolic regressor $S$ is an algorithm which takes a set of $n$ input-output pairs $\{(x_i, y_i)\}_{i=1}^n \sim \mathcal{X}\times\mathcal{Y}$  as input and returns a symbolic equation $e$ representing a function $f_e$ such that: $y\approx f_e(x)$,  $\forall (x,y)\in\mathcal{X}\times\mathcal{Y}$.
In this section, we describe our framework to \textit{learn} a parametrized symbolic regressor $S_\theta$ from a large number of training data.

\vspace{-0.40mm}
\subsection{Pre-training}
We pre-train a Transformer on hundreds of millions of equations which are procedurally generated for every minibatch.
As equations and datapoints can be generated quickly and in any amount using a computer and standard math libraries, we can train the network end-to-end to predict the equations on a dataset that is potentially unbounded.
We describe the exact process we use to generate the dataset in Section \ref{sec:experiments}.\\
An illustration of the main steps involved in the pre-training phase is shown in Fig. \ref{procedure}.

\vspace{-0.40mm}
\paragraph{Data}
During the pre-training phase, each training example consists of a symbolic equation $e$ which represents a function  $f_e: \mathbb{R}^{d_x} \rightarrow \mathbb{R}^{d_y}$, a set of $n$ input points $X = \{x_i\}_{i=1}^n$ and corresponding outputs $Y = \{f_e(x_i)\}_{i=1}^n$.
The distribution, $\mathcal{P}_{e, X}$, from which $e$ and the inputs $X$ are sampled will determine the inductive bias of the trained symbolic regressor and should be chosen to resemble the application domain.
In particular, $X$ can vary in size (i.e. $n$ is not fixed), and the individual inputs $x_i$ do not have to to be \textit{i.i.d} -- neither within $X$ nor across examples or batches.
For example, $\mathcal{P}_{e, X}$ could be polynomials of degree up to 6, and input sets of up to 100 points sampled uniformly from the range [0, 1].
In our experiments, an equation $e$ is represented by a sequence of symbols in prefix notation.
An equation $e$ can contain numerical constants that are re-sampled at each batch to increase the diversity of the data seen by the model.
In Section \ref{sec:experiments}, we describe the details of the data generation process we used in our experiments.
\vspace{-0.40mm}
\paragraph{Pre-training}
We train a parametric set-to-sequence model $S_\theta$ to predict the equation $e$ from the set of input-output points $X, Y$.
In our implementation, $S_\theta$ consists of an encoder and a decoder.
The encoder maps the $(x, y)$ sequence pairs for each equation into a latent space, resulting in a fixed-size latent representation $z$.
A decoder generates a sequence $\bar{e}$ given $z$: it produces a probability distribution $P(\bar{e}_{k+1} | \bar{e}_{1:k}, z)$ over each symbol, given the previous symbols and $z$.
The alphabet of $\bar{e}$ is identical to the one used for the original equations $e$, with one exception: 
unlike $e$, $\bar{e}$ does not contain any numerical constants.
Instead, it contains a special placeholder symbol `$\diamond$' which denotes the presence of a constant which will be fit at a later stage.
For example, if $e = 4.2\sin(0.3x_1) + x_2$, then $\bar{e} = \diamond\sin(\diamond x_1) + x_2$.
We refer to the equation where numerical constants are replaced by placeholders as the ``skeleton'' of the equation, and use the notation $\bar{e}$ to refer to the symbolic equation that replaces numerical constants with `$\diamond$'.
The model is trained to reduce the average loss between the predicted $\hat{e}$ and $\mathrm{skeleton}(e)$, i.e. the skeleton of the original equation.
Training is performed with mini-batches of $B$ equations each. The overall pre-training algorithm is reported in Algorithm~\ref{alg::alg1}.

\vspace{-2mm}

\begin{algorithm}[H]
\caption{Neural Symbolic Regression pre-training}\label{alg::alg1}
\begin{algorithmic} 
\REQUIRE $S_\theta$, batch size $B$, training distribution $\mathcal{P}_{e, X}$
\WHILE{not timeout}
\STATE $L \leftarrow 0$
\FOR{$i$ in $\{1..B\}$}
\STATE $e, X \leftarrow$ sample an equation and input set from $\mathcal{P}_{e, X}$
\STATE $Y \leftarrow \{f_e(x) |~ x \in X\}$
\STATE $\bar{e} \leftarrow \mathrm{skeleton}(e)$
\STATE $L \leftarrow L - \sum_k \log P_{S_\theta}(\bar{e}_{k+1} | \bar{e}_{1:k}, X, Y)$ 
\ENDFOR
\STATE Compute the gradient $\nabla_\theta L$ and use it to update $\theta$.
\ENDWHILE
\end{algorithmic}
\end{algorithm}

\vspace{-6mm}
\subsection{Test time}
\vspace{-0.50mm}

At test time, given a set of input-output pairs $\{(x_i, y_i)\}_i$ we encode them using the encoder into a latent vector $z$. From $z$ we iteratively sample candidates skeletons of symbolic equations $\hat{\bar{e}}$ from the decoder.
Finally, for each candidate, we fit the numerical constants $\diamond$ by treating each occurrence as an independent parameter.
This can be achieved using a non-linear optimizer, either gradient-based or black-box, by minimizing a loss between the resulting equation applied to the inputs and the targets $Y$.
In our experiments, we used beam-search to sample high-likelihood equation candidates from the decoder, and, like \citet{brenden}, BFGS \cite{practicalMethodsOfOptimization} on the mean squared error to fit the constants.
\vspace{-0.50mm}

\section{Experimental Set-up}\label{sec:experiments}

Here, we present the instantiation of the framework described in Section \ref{sec:method} that we evaluate empirically, and detail the baselines and datasets used to test it.
For the rest of the paper, we will refer to our implementation as NeSymReS\footnote{For \textbf{Ne}ural \textbf{Sym}bolic \textbf{Re}gression that \textbf{S}cales}.

\subsection{The Model $S_\theta$}
For the encoder we opted for the Set Transformer architecture from \citet{settransformer}, using the original publicly available implementation.\footnote{\href{https://github.com/juho-lee/set_transformer}{https://github.com/juho-lee/set$\textunderscore$transformer}}
We preferred this to the standard Transformer encoder, as the number $n$ of input-output pairs can grow to large values, and the computation in Set Transformers scales as $\mathcal{O}(nm)$ instead of $\mathcal{O}(n^2)$, where $m\ll{n}$ is a set of learnable inducing points \cite{inducing1, inducing2} we keep constant at $m=50$.
For the decoder we opted for a regular Transformer decoder \cite{transformer}, using the default PyTorch implementation.
Encoder and decoder have 11 and 13 million parameters respectively.
The hyperparameters chosen for both networks --- detailed in Section \ref{app:hyperparams} --- \textit{were not} fine-tuned for maximum performance.

\vspace{2ex}
\subsection{Pre-training Data Generator}\label{subsection:pretraining}
We sample expressions following the framework introduced in \cite{facebook}.
A mathematical expression is regarded as a unary-binary tree where nodes are operators and leaves are independent variables or constants. 
Once an expression is sampled, it is simplified using the rules built in the symbolic manipulation library SymPy \cite{10.7717/peerj-cs.103}.
This sampling method allows us to precisely constrain the search space by controlling the depth of the trees and the set of admissible operators, along with their prior probability of occurring in the generated expression.
We opted for scalar functions of up to three independent input variables (i.e. $d_x = 3$ and $d_y = 1$).
For convenience, we pre-sampled 10 million skeletons of equations with up to three numerical constants each.
At training time, we sample mini-batches of size $B =150$ of the following elements:
\begin{description}[topsep=0pt,itemsep=-1ex,partopsep=1ex,parsep=1ex,font=\normalfont]
    \item[\textit{Equation skeletons}] with constant placeholders placed randomly inside the expressions.
    \item[\textit{Constants values}] $C_1,C_2,C_3$, each independently sampled from a uniform distribution $\mathcal{U}(1,5)$. 
    \item[\textit{Support extrema}] $S_{1,j},S_{2,j}$, with $S_{1,j}<S_{2,j}$ uniformly sampled from $\mathcal{U}(-10,10)$ independently for each dimension $j = 1,\dots, {d_x}$.
    \item[\textit{Input points}] for each input dimension $j= 1,\dots, {d_x}$. A set of $n$ input points, $X_j = \{x_{i,j}\}_{i=1}^n$, is uniformly sampled from $\mathcal{U}(S_{1,j}, S_{2,j},n)$ .
\end{description}
We then evaluate the equations on the input points $X = \{x_{i}\}_{i=1}^n$ to obtain the corresponding outputs $Y$.

As $Y$ can take very large or very small values, this can result in numerical instabilities and exploding or vanishing gradients during training.
Therefore, we convert every $x_i$ and $y_i$ from float to a multi-hot bit representation according to the half-precision IEEE-754 standard. Furthermore, in order to avoid invalid operations (i.e dividing by zero, or taking the logarithm of negative values), we drop out input-output pairs containing NaNs.

We train the encoder and decoder jointly to minimize the cross-entropy loss between the ground truth skeleton and the skeleton predicted by the decoder as a regular language model.
We use Adam with a learning rate of $10^{-4}$, no schedules, and train for $1.5M$ steps. Overall, this results in about $225M$ distinct equations seen during pre-training.
See Appendix \ref{app:exp} for more details about training and resulting training curves.

\subsection{Symbolic Regression at Test Time}
Given a set of input-output pairs from an unknown equation $e$, we feed the points into the encoder and use beam-search to sample candidate skeletons from the decoder.
We then use BFGS to recover the values of the constants, by minimizing the squared loss between the original outputs and the output from the predicted equations.
Our default parameters at test time are beam-size 32, with 4 restarts of BFGS per equation.
We select the best equation from the set of resulting candidates based on the in-sample loss with a small penalty of 1e-14 per token of the skeleton.\footnote{While we found this strategy to work well in practice, a validation set for model selection might offer better performances with noisy data.}

\subsection{Evaluation}\label{ref:evaluation}

We evaluate our trained model on five datasets. Unless otherwise specified, for all equations we sample 128 points at test time.

\vspace{-0.35cm}

\paragraph{AI-Feynman (AIF)}
First, we consider all the equations with up to 3 independent variables from the AI-Feynman (AIF) database \cite{AIf1} \footnote{\href{https://space.mit.edu/home/tegmark/aifeynman.html}{https://space.mit.edu/home/tegmark/aifeynman.html}}. 
The resulting dataset consists of 52 equations extracted from the popular \textit{Feynman Lectures on Physics} series. We checked our pre-training dataset, and amongst the 10 million equation skeletons, all equations from AIF appear.
However, as mentioned in the previous subsection, the support on which they are evaluated, along with the constants and number of points per equation, is continuously sampled at every training iteration, making it impossible to exactly see any of the test data at training time.

\paragraph{Unseen Skeletons (SOOSE)}
This dataset of 200 equations is specifically constructed to have zero overlap with the pre-training set, meaning that its equations are all symbolically and numerically different from those included in the pre-training set.
We call it SOOSE, for strictly out-of-sample equations.
Compared to AIF, these equations are on average significantly longer and more complex (see Table \ref{tab:SOOBE}).
The sampling distribution for the skeletons is the same as the pre-training distribution, but we instantiate three different versions: with up to three constants (same as pre-training distribution, SOOSE-WC); no constants (SOOSE-NC); constants everywhere (SOOSE-FC, for full constants), i.e. one constant term for each term in the equation.
The latter is extremely challenging, and since NeSymReS was only pre-trained with up to three constants, it is far from its pre-training distribution.

\paragraph{Nguyen Dataset}
This dataset consists of 12 simple equations \textit{without} constants beyond the scalars 1 and 2, each with up to 2 independent variables.
Nguyen was the main benchmark used in \cite{brenden}.
There are terms that appear in three ground truth equations that are \textit{not} included in the set of equations that our model can fit, specifically $x^6$, and $x^y$, which therefore caps the maximum accuracy that can be reached by our model on this dataset.

\begin{figure*}[t]
    \centering
    \includegraphics[width=0.98\linewidth]{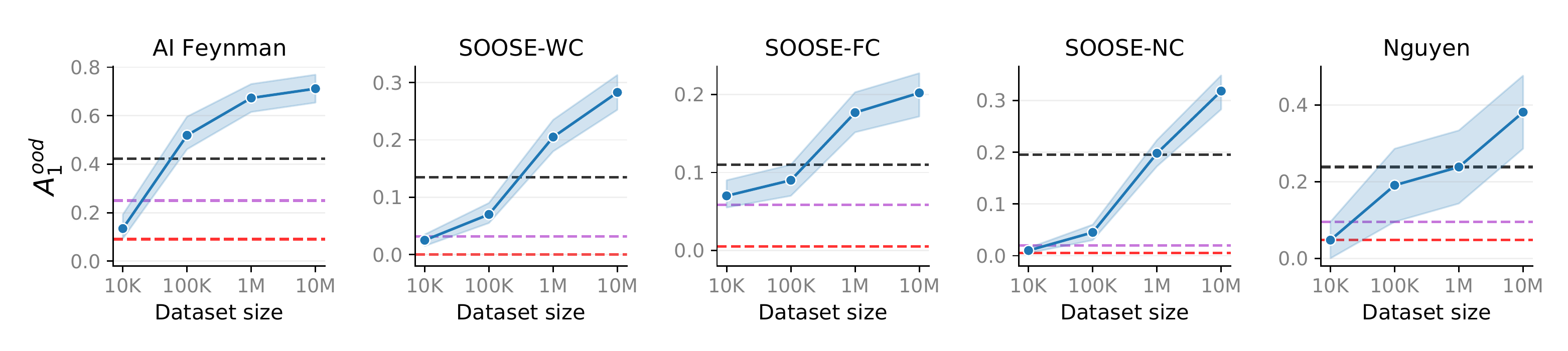}
    \includegraphics[width=0.6\linewidth]{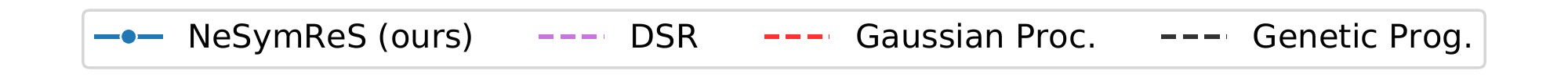}
    \caption{Accuracy as a function of the size of the pre-training dataset, for a fixed computational budget ($\sim$100 s) at test time. We report reference values for the baselines to emphasize that these approaches do not improve with experience over time.}
        
    \label{fig:training_amounts}
\end{figure*}

\vspace{-0.50mm}
\subsection{Baselines}
We compare the performance of our method with the following baselines:

\vspace{-0.50mm}
\begin{description}
    \vspace{-0.50mm}
    \item[Deep Symbolic Regression (DSR)] \cite{brenden} Recently proposed RNN-based reinforcement learning search strategy for symbolic regression. We use the open-source implementation provided by the authors\footnote{\href{https://github.com/brendenpetersen/deep-symbolic-regression}{https://github.com/brendenpetersen/deep-symbolic-regression}}, with the setting that includes the estimation of numerical constants in the final predicted equation.
    \vspace{-0.50mm}
    \item[Genetic Programming] \cite{gp2} Standard GP-based symbolic regression based on the open-source Python library \texttt{gplearn} \footnote{\href{https://gplearn.readthedocs.io/en/stable/}{https://gplearn.readthedocs.io/en/stable/}}.
    \vspace{-0.50mm}
    \item[Gaussian Processes] \cite{GProcess} Standard Gaussian Process regression with RBF and constant kernel. 
    We use the open source \texttt{sklearn} implementation\footnote{\href{https://scikit-learn.org/stable/modules/gaussian_process.html}{https://scikit-learn.org/stable/modules/gaussian$\textunderscore$process.html}}.   
\end{description}

All details about baselines are reported in Appendix \ref{app:hyperparams}.

Two notable exclusions are AIF \cite{AIf1} and EQL \cite{martius2016extrapolation}. 
As also noted by \citet{brenden}, in cases where real numerical constants are present or the equations are not separable, the former still requires a complementary symbolic regression method to cope with the discrete search. 
The latter lacks too many basis functions that appear in the datasets we consider, preventing it from recovering most of the equations. Moreover, its average runtime and number of points required to solve the equations indicated in \cite{martius2016extrapolation, SahooLampertMartius2018:EQLDiv} are three orders of magnitudes higher than the standards reported by the aforementioned baselines.

\subsection{Metrics}\label{subsec:Metrics}
Evaluating whether two equations are equivalent is a challenging task in the presence of real valued constants.

We distinguish between accuracy \textit{within} the training support ($A^{\text{iid}}$), and outside of the training support ($A^{\text{ood}}$).
$A^{\text{iid}}$ is computed with 10k points sampled uniformly in the training support.
$A^{\text{ood}}$ is computed with 10k points in an extended support as detailed in Appendix \ref{app:exp}, and it will be the main metric of interest.

We further distinguish between two metrics, accuracy $A_1$ and accuracy $A_2$, each of which can be either computed \textit{iid} or \textit{ood}.
Accuracy $A_1$ is computed as follows: for every point $(x, y)$ and prediction $f_{\hat{e}}(x) = \hat{y}$, the point is correctly classified if \texttt{numpy.isclose($y$, $\hat{y}$)} returns True.\footnote{With parameters atol 1e-3 and rtol 0.05.}
Then, an equation is correctly predicted if $> 95\%$ of points are correctly classified. 
For this metric we can keep \textit{all} outputs, including NaNs and $\pm \infty$, which are still representative of whether the symbolic equation was identified correctly.
Accuracy $A_2$ is computed by measuring the coefficient of determination $R^2$ between $y$ and $\hat{y}$, excluding NaNs and $\pm \infty$.
An equation is correctly identified according to $A_2$ if the $R^2 >$ 0.95.
We found the two metrics to correlate significantly, and in the interest of clarity we will use only $A_1$ in the main text, and show results with $A_2$ in the Appendix \ref{app:res}.

\vspace{-1mm}
\section{Results}
\vspace{-1mm}
We test three different aspects of the proposed approach: (i) To what extent does performance improve as we increase the size of the pre-training data? (ii) How does our approach compare to state-of-the-art methods in symbolic regression? (iii) What is the impact of the number of input-output pairs available at \textit{test time}? 

\label{sec:results}

\begin{figure*}[t]
    \centering
    \includegraphics[width=0.99\linewidth]{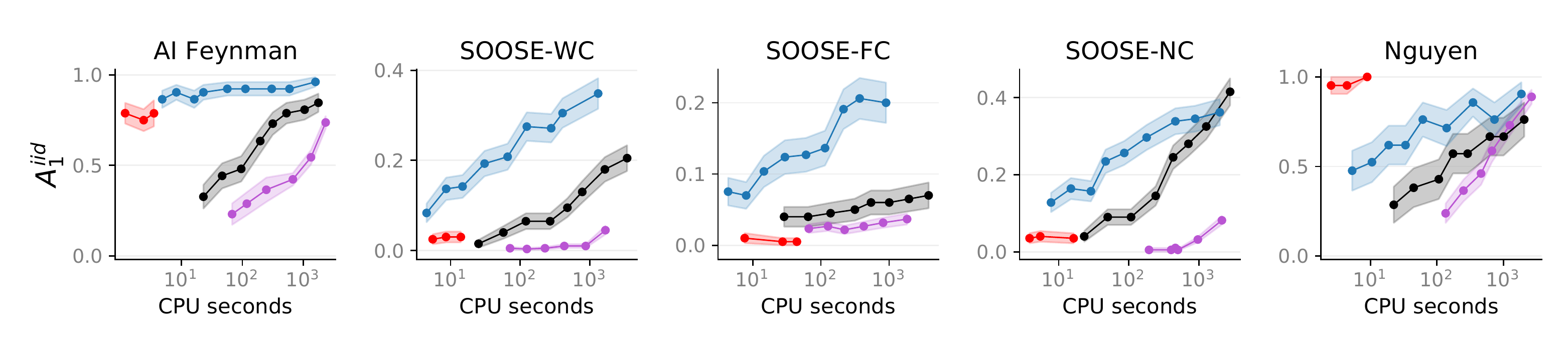}
    \\
    \includegraphics[width=0.65\linewidth]{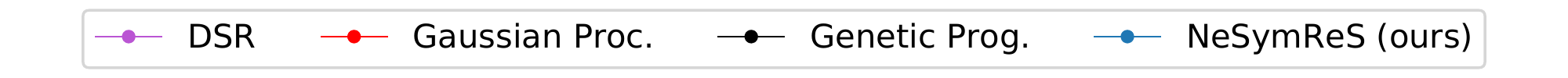}
    \vspace{-2mm}
    \caption{Accuracy in distribution as a function of time for all methods ran on a single CPU per equation.}
    \label{fig:acc_vs_time_iid} %
\end{figure*}

\vspace{-4mm}
\begin{figure*}[t]
    \centering
    \includegraphics[width=0.99\linewidth]{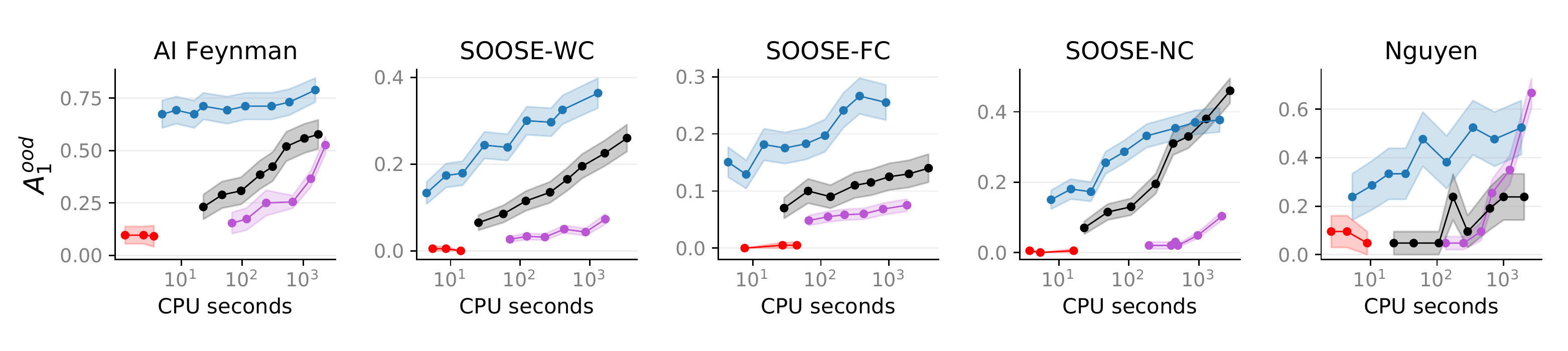}%
    \\
    \includegraphics[width=0.65\linewidth]{figs/plots_2021_june_camera_ready/acc_vs_time/acc_vs_time_5cols_legend.pdf}
    \vspace{-2mm}
    \caption{Accuracy out of distribution as a function of time for all methods ran on a single CPU per equation.}
    \label{fig:acc_vs_time_ood}
\end{figure*}

\subsection*{(i) Accuracy as a Function of Pre-training Data}
\vspace{-0.8mm}
In order to test the effect of pre-training data on test performance, we trained our NeSymReS model on increasingly larger datasets. More specifically, we consider datasets consisting of 10K, 100K, 1M and 10M equation skeletons.
Every aspect of training is the same as described in Section \ref{sec:experiments}.
We train all models for the same number of iterations, but use early stopping on a held-out validation set to prevent overfitting.

In Figure \ref{fig:training_amounts} we report the accuracy on the 5 test sets using a beam size of 32 for NeSymReS, and for all baselines whatever hyperparameter configuration that used comparable (but strictly no less) amount of computing time.
In all datasets, increasing the size of the pre-training data results in higher accuracy for NeSymReS.
Note that the baselines do not make use of the available pre-training data, and as such it does not have any effect on the performance at test time.
From here onwards, we will always use the model pre-trained on 10M equation skeletons.

\vspace{-2mm}
\textit{Conclusion:} 
The performance of NeSymReS steadily improves as the size of the pre-training dataset increases, exploiting the feature that symbolic equations can be generated and evaluated extremely quickly and reliably with computers. 
The trend observed appears to continue for even larger datasets, in accordance to \cite{scaling}, which leaves open interesting avenues for extremely large scale experiments.

\subsection*{(ii) Accuracy as a Function of Test-time Compute.}
For every method (including baselines), we vary the corresponding hyper-parameter that increases how much time and compute is invested at test time to recover an equation from observing a fixed set of input-output pairs.
We report the hyper-parameters and ranges in Table \ref{tab:acc_time}.

Making a fair comparison of run-times between different methods is another challenging task.
To make the comparison as fair as possible, we decided to run every method on a single CPU at the time.
Note that this is clearly a sub-optimal hardware setting for our 26-million parameters Transformer, which would be highly parallelizable on GPU.

The results on all five datasets are shown in Figure \ref{fig:acc_vs_time_iid} and Figure \ref{fig:acc_vs_time_ood}.
On all datasets, our method outperforms all baselines both in time and accuracy by a large margin on most budgets of compute.
On AIF our NeSymRes is more than three orders of magnitudes faster at reaching the same maximum accuracy as the second-best method, i.e. Genetic Programming, despite running on CPU only.
We attribute the low accuracy achieved by \cite{brenden} to the presence of constants, to the fact that their model does not directly observe the input-output pairs, and the use of REINFORCE \cite{Williams2004SimpleSG}.
The Gaussian Process baseline performs extremely well in distribution, reaching high accuracy in a very short amount of time, but poorly out of distribution.
This is expected as it does not try to regress the symbolic equation.
On Nguyen, NeSymReS achieves relatively high scores more rapidly than the other baselines. For large computation times ($\approx 10^3$ seconds) NeSymReS performs comparably with DSR despite the latter being fine-tuned on two equations of the benchmark (Nguyen-7 and Nguyen-10).
The relatively lower performance of NeSymReS on SOOSE-NC can be explained by the fact that both datasets do not have any constants in the equations, while NeSymReS is trained with a large prior on the presence of constants.

\begin{table}[b]
\centering
\caption{Hyper-parameters that vary to increase the amount of compute invested by every method.}
\small
\label{tab:acc_time}
\begin{tabular}{@{}lll@{}}
\toprule
Method                         & Hyper-param  & Range \\
\midrule
G. Proc. \cite{GProcess}       & Opt. restarts             & $\{8, 16, 32\}$  \\
Genetic Prog. \cite{gp2}   & Pop. size & $\{2^{10}, ..., 2^{17}\}$ \\
DSR \cite{brenden} & Epochs   & $\{2^2, ..., 2^7\}$   \\
NeSymReS (ours)                & Beam size       & $\{2^0,..., 2^8\}$ \\
\bottomrule

\end{tabular}
\end{table}

\textit{Conclusion:} By amortizing the computation performed at pre-training time, NeSymReS is extremely accurate and efficient at test time, even running on CPU.

\subsection*{(iii) Performance Improves with more Points $p$}

\begin{figure}[H]
    \centering
    \includegraphics[width=0.49\linewidth]{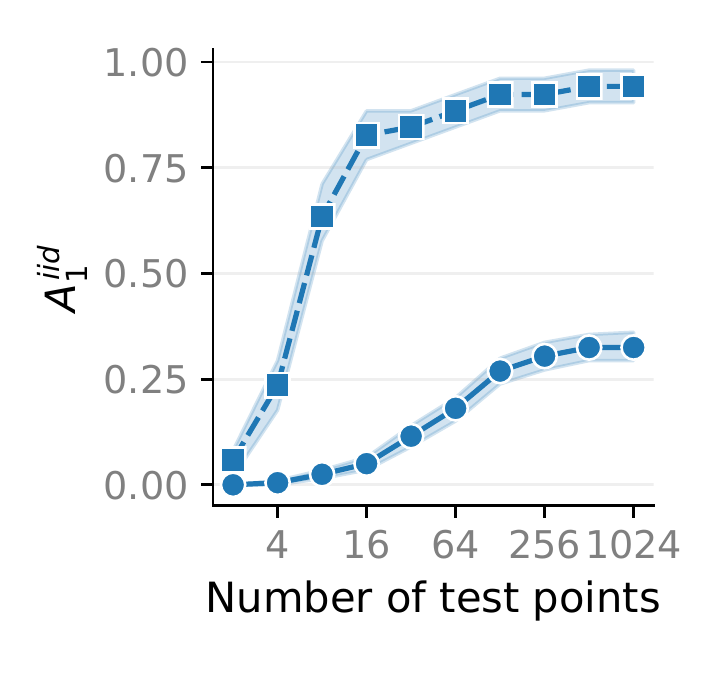}
        \includegraphics[width=0.49\linewidth]{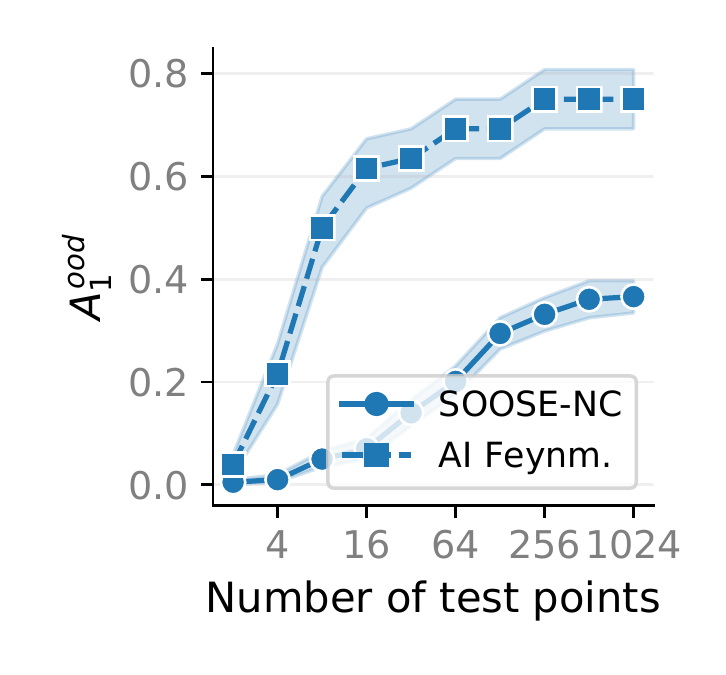}
    \vspace{-5mm}
        \caption{Accuracy as a function of number of input-output pairs observed at test time.%
        }
    \label{fig:acc_vs_points}
\end{figure}

In practice, depending on the context, a variable number of input-output pairs might be available at test time.
In Figure \ref{fig:acc_vs_points}, we report the accuracy achieved for a number of input-output points that varies in the range from 1 to 1024.
Even though NeSymReS was pre-trained with no more than 500 points, it still performs reliably with fewer points.

\textit{Conclusion:} NeSymReS is a flexible method and its performance is robust to different numbers of test data, even when such numbers differ significantly from those usually seen during pre-training. Furthermore, its accuracy levels grow with the number of points observed at test time. 

\vspace{-3mm}

\section{Discussion}\label{sec:conclusion}
Building on the recent successes of large scale pre-training, we have proposed the first method that \textit{learns the task} of symbolic regression.
This approach deviates from the majority of existing techniques in the literature which need to be retrained from scratch on each new equation and does not improve over time with access to data and compute \cite{bitterlesson}. 
We showed empirically that by pre-training on a large distribution of millions of equations, this simple approach outperforms several strong baselines, and that its performance can be improved by merely increasing the size of the dataset.
The key feature that enables this approach is that --- unlike for computer vision and natural language --- high-quality training data can be generated efficiently and indefinitely using any standard math library and a computer.

In pre-training, the data generation plays a crucial role within our framework.
By changing this distribution over equations (including support, constants, number of terms and their interactions), it is possible for the user to finely tune the inductive bias of the model, adapting it to specific applications.
In light of its favourable scaling properties and its powerful prior over symbolic expression, we believe that our model could find applications in several domains in the Natural Sciences and engineering, control, and model-based Reinforcement Learning. 
The scale of our experiments is still relatively small compared to the largest large-scale experiments run to date \cite{gpt3,bert, simclr2}, both in terms of dataset and model sizes.
Nonetheless, the results we showed already seem to indicate that NeSymReS could improve significantly with access to extremely large scale compute. 

\paragraph{Time and Space Complexities}
The approach we presented scales favorably over several dimensions: computation scales linearly in the number of input-output points due to the Set Transformer \cite{settransformer}, and linearly in the number of input dimensions.
For future work, it would be interesting to train even larger models on larger datasets with more than three independent variables.

\paragraph{Limitations} 
Even though our approach can scale to an arbitrary number of input and output dimensions, there are limitations that should be considered.
Fitting the constants using a non-linear optimizers like BFGS can prove to be hard if the function to be optimized has several local minima.
In this case, other optimization strategies that can deal with non-convex loss surfaces might be beneficial, such as CMA-ES \cite{CMA-ES}.
One more limitation of our approach is that the pre-trained model as presented cannot be used at test time if the number of input variables is larger than the maximum number of variables seen during pre-training.
Finally, one more limitation of the neural network we adopt is that it does not directly interact with the function evaluator available in the math libraries of most computers.
If, for example, the first candidate sampled from the network is completely wrong, our current approach cannot adjust its posterior over equations based on this new evidence, but simply sample again.

\paragraph{Conclusions} 
What are the desirable properties of a strong symbolic regressor? It should:
\begin{itemize}
    \item \textit{scale} favourably with the number of datapoints observed at test time and with the number of input variables;
    \item \textit{improve over time} with experience;
    \item \textit{be targetable} to specific distributions of symbolic equations;
    \item \textit{be flexible} to accommodate very large or very small values.
\end{itemize}
In this paper, we showed that all of these properties can be obtained, and provided a simple algorithm to achieve them in the context of symbolic regression.
Our largest pre-trained model can be accessed on our repository. %

\section*{Acknowledgements}
We thank Guillame Lample for the helpful discussion, and Riccardo Barbano for proofreading the manuscript.
LB acknowledges the CSEM Data Program Fund for funding.
GP acknowledges the Max Planck ETH Center for Learning Systems for funding.
AN acknowledges the International Max Planck Research School for Intelligent Systems for funding. 
This work was also supported by the German Federal Ministry of Education and Research (BMBF) through the Tübingen AI Center (FKZ: 01IS18039B), and the ML cluster funded by the Deutsche Forschungsgemeinschaft (DFG, German Research Foundation) under Germany's Excellence Strategy - EXC number 2064/1 - Project number 390727645.

\bibliography{bibliography}
\bibliographystyle{icml2021}

\newpage

\appendix
\onecolumn
\section{Model details}
\label{app:hyperparams}
\subsection{NeSymReS Transformer Details}
The model consists of an encoder and a decoder. The encoder takes as input numerical data, $X\in\mathbb{R}^{(d_x+d_y)\times{n}}$, where $d_x$ is the number of independent variables, $d_y$ the number of dependent variables and $n$ the number of support points. In order to prevent exploding gradient and numerical instabilities, we convert each entry of $X$ into a multi-hot bit representation according to the half precision IEEE-754 standard. This operation yields a new input tensor, $\Tilde{X}\in\mathbb{R}^{(d_x+d_y)\times{b}\times{n}}$ where $b=16$ is the dimension of the bit representation. The output of the encoder is a latent vector $z\in\mathbb{R}^{d_z}$, providing a compressed representation of the equation to be modelled. Such latent vector is then used to condition the decoder via a standard Transformer multi-head attention mechanism. During the pre-training phase, the input of the decoder is given by the sequence of tokens representing the ground truth-equation expressed in prefix notation. Such sequence is opportunely masked in order to prevent information leakage in the decoder forward step. The output is then given by a string of symbols, representing the predicted equation, again in prefix notation. During inference, the decoder is only provided with the information from the latent vector $z$ and generates a prediction autoregressively. We use \textit{beam search} to obtain candidate solutions. After removing potentially invalid equations, the remaining equations are modified to include constant placeholders with the procedure described in Appendix \ref{app:exp}. 
Using BFGS, we fit these constants. 
We select the best equation among the so-found candidates, based on the validation loss, with an added regularization penalty of $10^{-14}$ for each token in the skeleton. 
Note that BFGS is currently the most time-consuming step of our pipeline. 
While in all our experiments we run the optimization procedure serially (i.e., one candidate equation at the time), the procedure can be easily parallelized across equations.

Encoder and decoder use the same hidden dimension $H$ and number of heads, $h$ for their multi-head attention modules. In the following, we provide further details about the architectural design choices, hyper-parameters and library of functions used by our model.
We trained our model on a single GeForce RTX 2080 GPU for $3$ days.

\paragraph{Encoder}
For the encoder, we opted for the Set Transformer architecture from \citet{settransformer}. Our choice is motivated in light of the better scaling properties of this method when it comes to input lenght $n$, i.e. $\mathcal{O}(nm)$ compared to the standard transformer encoder, $\mathcal{O}(n^2)$, where $m$ is a set of trainable inducing points. Referring to the notation of the original paper, our encoder is formed by $n_e$ Induced Set Attention Blocks (ISABs) and one final component performing Pooling by Multihead Attention (PMA). ISABs differ from the multi-head self attention blocks present in the original Transformer architecture, since they introduce $m<n$ learnable inducing points that reduce the computation burden associated with the self-attention operation. PMA allows us to aggregate the output of the encoder into $d_z$ trainable abstract features representing a compressed representation of the input equation. Overall, the encoder consists of $11M$ trainable parameters.
All the hyper-parameters of the encoder are listed in Table \ref{tab:enc}.

\paragraph{Decoder}
The decoder is a standard Transformer decoder. It has $n_d$ layers, hidden dimension $H$, $h$ attention heads. Input symbols are encoded into the corresponding token embeddings and information about relative and absolute positions of the tokens in the sequence is injected by adding learnable positional encodings to the input embeddings. 
Two different masks are used to avoid information leakage during the forward step and to make the padding token hidden to the attention modules. 
In total, our dictionary is formed by $s$ different tokens, including binary and unary operators and independent variables. A list of all the elements of our dictionary is provided in Table \ref{tab::toks}.
Overall, the decoder consists of $15M$ trainable parameters. All the hyper-parameters of the decoder are listed in Table \ref{tab:dec}.

\begin{table}[h]
\parbox{.45\linewidth}{
\centering
\caption{Encoder hyper-parameters.}
\small
\label{tab:enc}
\begin{tabular}{@{}lll@{}}
\toprule
Parameter name                         & Symbol  & Value \\
\midrule
Number of ISABs       & $n_e$   & 5  \\
Hidden dimension    &$H$ &512 \\
Number of heads & $h$   & 8  \\
Number of PMA features        & $d_z$      &10 \\
Number of inducing points        & $m$      &50 \\
\bottomrule

\end{tabular}
}
\hfill
\parbox{.45\linewidth}{
\centering
\caption{Decoder hyper-parameters.}
\small
\label{tab:dec}
\begin{tabular}{@{}lll@{}}
\toprule
Parameter name                         & Symbol  & Value \\
\midrule
Number of layers       & $n_d$   & 5  \\
Hidden dimension    &$H$ &512 \\
Number of heads & $h$   & 8  \\
Embedding dimension & $s$  &32\\
\bottomrule
\end{tabular}
}
\end{table}

\begin{table}[h]
\centering
\begin{tabular}{|c|c|}
\hline
\multicolumn{1}{|l|}{\textbf{Symbol}} & \multicolumn{1}{l|}{\textbf{Integer Id}}         \\ \hline\hline
sos                                   & 1                                                \\ \hline
eos                                   & 2                                                \\ \hline
$x$                                   & 3                                                \\ \hline
$y$                                   & 4 \\ \hline
$z$                                   & 5                                                \\ \hline
$c$                                   & 6                                                \\ \hline
$\arccos$                           & 7                                                \\ \hline
$+$                                   & 8                                                \\ \hline
\end{tabular}
\quad
\begin{tabular}{|c|c|}
\hline
\multicolumn{1}{|l|}{\textbf{Symbol}} & \multicolumn{1}{l|}{\textbf{Integer Id}}          \\ \hline\hline

$\arcsin$                           & 9                                                \\ \hline
$\arctan$                           & 10                                                \\ \hline
$\cos$                                     & 11                                                \\ \hline
$\cosh$                                     & 12                                                \\ \hline
$\coth$                                     & 13                                               \\ \hline
$\div$                                     & 14                                               \\ \hline
$\exp$                                     & 15                                               \\ \hline
$\ln$                                 & 16                                                \\
\hline
\end{tabular}
\quad
\begin{tabular}{|c|c|}
\hline
\multicolumn{1}{|l|}{\textbf{Symbol}} & \multicolumn{1}{l|}{\textbf{Integer Id}}          \\ \hline\hline

$\times$                                  & 17
        \\ \hline
Pow                                     & 18                                                \\ \hline
$\sin$                                     & 19                                                \\ \hline
$\sinh$                                     & 20                                                \\ \hline
$\sqrt{}$                                     & 21                                                \\ \hline
$\tan$                                     & 22                                                \\ \hline
$\tanh$                                     & 23                                                \\ \hline
$-3$                                     & 24                                                \\ \hline
\end{tabular}
\quad
\begin{tabular}{|c|c|}
\hline
\multicolumn{1}{|l|}{\textbf{Symbol}} & \multicolumn{1}{l|}{\textbf{Integer Id}}          \\ \hline\hline

$-2$                                     & 25                                                \\ \hline
$-1$                                     & 26                                                \\ \hline
$0$                                     & 27    
\\ \hline
$1$                                     & 28                                                \\ \hline
$2$                                     & 29                                                \\ \hline
$3$                                     & 30                                                \\ \hline
$4$                                     & 31    
\\ \hline
$5$                                     & 32    
\\ \hline
\end{tabular}
\caption{Symbols available for expression generation and their corresponding integer tokens. The symbols \enquote{sos} and \enquote{eos} stand for \enquote{start of sequence} and \enquote{end of sequence} respectively. The padding symbol is not reported in the table and is associated with the token $0$.
Note that \textbf{not all of these symbols} appear in our pre-training dataset, as detailed in Table \ref{tab:unormalized_prob} we only used a small subset for our experiments.
}
\label{tab::toks}
\end{table}

\subsection{Baselines}

\paragraph{Deep Symbolic Regression (DSR)}
For DSR, we use the standard hyper-parameters provided in the open-source implementation of the method, with the setting that includes the estimation of numerical constants in the final predicted equation. DSR depends on two main hyper-parameters, namely the entropy coefficient $\lambda_{H}$ and the risk factor $\epsilon$. The first is used to weight a bonus proportional to the entropy of the sampled expression which is added to the main objective. The second intervenes in the definition of the final objective with depends on the $(1-\epsilon)$ quantile of the distribution of rewards under the current policy. According with the open-source implementation and the results reported in \cite{brenden}, we choose $\epsilon= 0.05$ and $\lambda_{H}= 0.005$. The set of symbols available to the algorithm to form mathematical expressions is given by $\mathcal{L}= \{+,-,\times,\div,\sin,\cos,\exp,\ln,c\}$, where $c$ stands for the constant placeholder.

\paragraph{Genetic Programming}
For Genetic Programming, we opt for the open-source Python library \texttt{gplearn}. Our choices for the hyper-parameters are listed in Table \ref{tab::gp} and mostly reflect the default values indicated in the library documentation. The set of symbols available to the algorithm to form mathematical expressions is the default one and is given by $\mathcal{L}= \{+,-,\times,\div,\sqrt,\ln, \exp, \textit{neg},\textit{inv},\sin,\cos\}$, where $\textit{neg}$ and $\textit{inv}$ stand for \enquote{negation} ($x \mapsto -x$), and inversion ($x \mapsto x^{-1}$), respectively.

\begin{table}[h]
\centering
\caption{Genetic Programming hyper-parameters. The parameter \textit{Population size} is varied within the range indicated during the experiments reported in Section \ref{sec:results}.}
\small
\label{tab::gp}
\begin{tabular}{@{}lll@{}}
\toprule
Parameter name                         & Value \\
\midrule
Population size  &$\{2^{10}, ..., 2^{15}\}$\\
Selection type   &Tournament\\
Tournament size (k)     &20 \\
Mutation probability    &0.01  \\
Crossover probability   &0.9\\
Constants range  &$(-4\pi, 4\pi)$\\
\bottomrule
\end{tabular}
\end{table}

\paragraph{Gaussian Processes}
This is the only baseline that is \textit{not} a symbolic regression method per se, as it learns a mapping from $x$ to $y$ directly.
The appealing property of Gaussian Processes is that they are very accurate in distribution, and are very fast to fit in the regime we considered.
We opted for the open-source \texttt{sklearn} implementation of Gaussian Process regression with default hyper-parameters. The covariance is given by the product of a constant kernel and an RBF kernel. Diagonal Gaussian noise of variance $10^{-10}$ is added to ensure positive-definitness of the covariance matrix. L-BGFS-B is used for the optimization of the marginal likelihood with the number of restarts varied as indicated in Table \ref{tab:acc_time}.

\paragraph{A note about function sets}
Unfortunately, not all methods support all primitive functions that appear in a given dataset.
For example, NeSymReS \textit{could} support $x^6$, and $x^y$ --- that appear in the Nguyen dataset described in Appendix \ref{app:res} --- but as we did not include these primitives in the pre-training phase, the version we use in our experiments will not be able to correctly recover these equations.
DSR and the implementation of Genetic Programming that we adopted are both lacking $\arcsin$ in their function set.

While missing primitives lowers the upper bound in performance that a method can reach for a given dataset, it also makes it easier to fit the other equations that \textit{do not} contain those primitives, as the function set to search is effectively smaller.

\section{Experimental details}
\label{app:exp}

\subsection{Training}

\begin{table}[h]
\centering
\begin{tabular}{l|llllllllllll}
\toprule
Operator         &  $+$ &  $\times$ &  $-$ &  $\div$ &  $\sqrt{}$ &  Pow &  $\ln$ &  $\exp$ &  $\sin$ &  $\cos$ &  $\tan$ &  $\arcsin$ \\
Unormalized Prob &   10 &   10 &    5 &    5 &     4 &    4 &   4 &    4 &    4 &    4 &    4 &     1 \\
\bottomrule
\end{tabular}

\caption{Operators and their corresponding un-normalized probabilities of being sampled as parent node.}
\label{tab:unormalized_prob}
\end{table}%
\paragraph{Training Dataset Generation}
For generating skeletons, we built on top of the method and code proposed in \cite{facebook}, which samples expression trees. 
For our experiments, each randomly-generated expression tree has 5 or fewer non-leaf nodes. We sample each non-leaf node following the unnormalized weighted distribution shown in Table \ref{tab:unormalized_prob}. 
Each leaf node has a probability of 0.8 of being an independent variable and 0.2 of being an integer. Trees that contain the independent variable $x_2$ must also have the independent variable $x_1$. Those containing the independent variable $x_3$ must also include the independent variables $x_1$ and $x_2$.
We then traverse the tree in pre-order and obtain a semantically equivalent string of the expression tree in a prefix notation. We convert the string from prefix to infix notation and simplify the mathematical expression using the Sympy library. 
The resulting expression is then modified to include constant placeholders as explained in the following paragraph.
This expression is what we refer to as a \textit{skeleton}, as the value of constants has not been determined yet.

For our experiments, we repeat the procedure described above to obtain a pre-compiled dataset of 10M equations.
To compile the symbolic equation into a function that can be evaluated by the computer on a given set of input points, we relied on the function \textit{lambidfy} from the library Sympy.
We store the equations as functions, in order to allow for the support points and values of the constants to be resampled at mini-batch time during pre-training.
We opted for a partially pre-generated dataset instead of sampling new equations for every batch in order to speed up the generation of training data for the mini-batches.

\begin{figure*}[h]
    \centering
     \includegraphics[width=0.99\linewidth]{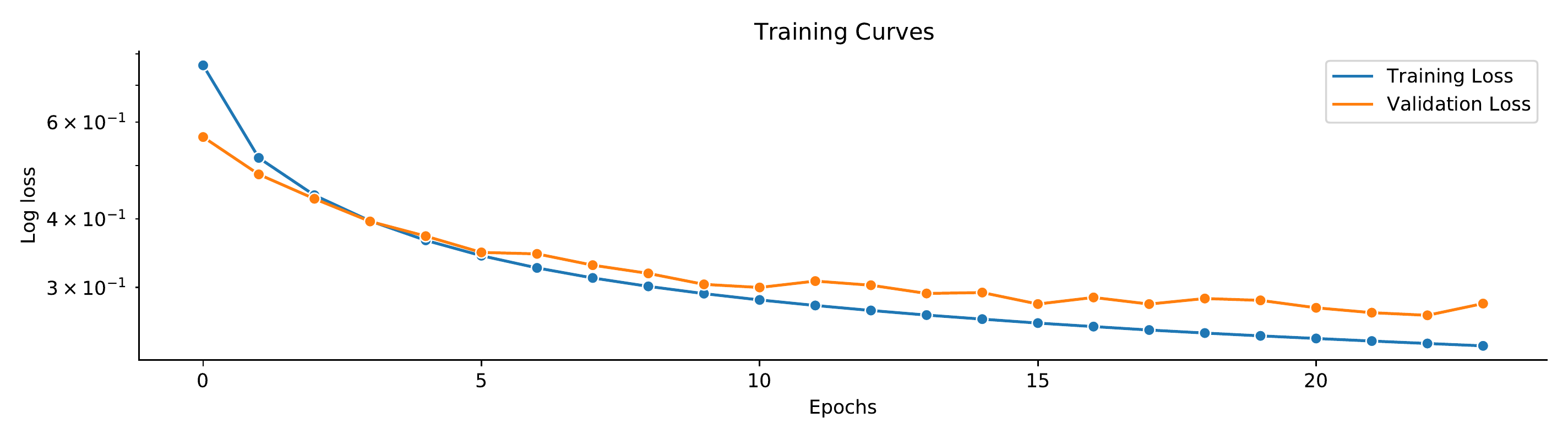}  \\
    \caption{Loss as a function of pre-training for NeSymReS}\label{training_curve}
\end{figure*}

\paragraph{Training Details}
As described in Section, \ref{subsection:pretraining}, during training we sampled mini-batches of size $B =150$ from the generated dataset.
For each equation, we first choose the number of constants, $n_c$, that differ from one. $n_c$ is randomly sampled within the interval ranging from $0$ to $\min(3,N_c)$ where $N_c$ is the maximum number of constants than can be placed in the expression. Then, we sample the constants' values from the uniform distribution $\mathcal{U}(1,5)$. We randomly select $n_c$ among the available placeholders and replace them with the previously obtained numerical values. The remaining constants are set to one. We then generate up to $500$ support points by sampling- independently for each dimension - from uniform distributions with varying extrema as described in Section \ref{sec:experiments}. If the equation does not contain a given independent variable, such variable is set to $0$ for all the support points. For convenience, we drop input-output pairs containing NaNs and entries with an absolute value of $y$ above $1000$.
Finally we take the equation in the mini-batch with the minimum number of points, and drop valid points from the other equations so that the batch tensor has a consistent length across equations.
Figure \ref{training_curve} shows the training and validation curves of the main model used for all our experiments. 

\paragraph{Training Dataset Distribution}
The dataset does not consist of unique mathematical expressions. Indeed, some skeletons are repeated, and some skeletons are mathematically equivalent. Overall, within the 10M dataset, we have $\sim{1.2}$M unique skeletons. Since longer expressions tend to be simplified into shorter expressions during the dataset generation, shorter expressions are the most frequent ones. 
We report in Table \ref{tab:frequency} the ten most recurrent expressions.
Of these $1.2$M unique skeletons, at least $96.3 \%$ represents unique (numerically distinct) mathematical expressions. The counting procedure is described in the next paragraph. 
The average length of an expression in infix notation is 8.2 tokens. The minimum and the maximum are 1 and 23 respectively, which corresponds in infix notation to the expressions $x$ and $\frac{x^{2} \operatorname{asin}^{2}{\left(x \right)}}{- x_1^{2} \operatorname{asin}^{2}{\left(x \right)} + 1}$.

\begin{table}[h]
\centering
\begin{tabular}{l|llllllllll}
\toprule
Expression &    $x_1$ &  $x_1 + x_2$ &  $x_2^{2}$ &  $x_1 x_2$ &  $x_1 + x_2 + x_3$ &  $x_1 + 1$ &  $- x_1$ &  $x_1 - 1$ &  $e^{x_1}$ &  $\cos{\left(x_1 \right)}$ \\
Frequency  &  18649 &     6716 &     4789 &   4707 &         3781 &     3727 &   3698 &     2917 &     2594 &                     2157 \\
\bottomrule
\end{tabular}

\caption{10 most frequent expressions in the dataset}
\label{tab:frequency}
\end{table}%

\paragraph{Addition of Numerical Constants}
\label{para:addition_constants}
During dataset generation and inference, we introduce constant placeholders by attaching them to the generated or predicted skeletons. This step is carried out by multiplying all unary operators in the expressions by constant placeholders (except for the \enquote{pow} operator). The same procedure is repeated with independent variables for which also additive constants are introduced. Longer expressions tend to have more placeholders in comparison to shorter ones.

\subsection{Evaluation details}

All results reported, i.e. for all methods and datasets, are accuracies over all equations in the dataset. Error bars in all plots denote the standard error of the mean estimate.

\paragraph{Metrics Details}
As detailed in \ref{subsec:Metrics} we evaluate performances both within the training support ($A^{\text{iid}}$) and outside of the training support ($A^{\text{ood}}$). More specifically, for the latter the support is created as follows: given an equation with in-sample support of $(lo, hi)$, we extend the support of each side by $(hi-lo)$ for every variable present in the equation. 

\paragraph{Creating the SOOBE Dataset}
The SOOBE (stricly out-of-sample equations) dataset contains entirely different skeletons from the pre-training dataset, which do not overlap numerically nor symbolically. To create it, we list all the different expressions in the training dataset and then randomly sample from this set, excluding the sampled expressions from the training set.
We first sample a random support of 500 points from the uniform distribution $\mathcal{U}(-10,10)$, for each independent variable. %
Two expressions are different if their images, given the support points, are different. 
Note that this is a conservative criterion, as two expressions may have the same image in the sampled support (relatively to a fixed tolerance), yet being distinct.

\paragraph{Benchmarks}
As explained in \ref{ref:evaluation} we evalutate our trained model on five datasets: AI-Feynman, SOOBE-WC, SOOBE-NC, SOOBE-FC, Nguyen. All the equations of AI-Feynman used in our evaluation are listed in table \ref{tab:ai_fey}. 50 randomly sampled equations out of 200 from the SOOBE dataset, are listed in table \ref{tab:SOOBE}.

\begin{table}[h]
\centering
\begingroup
\renewcommand{\arraystretch}{2} %
\resizebox{\textwidth}{!}{
\begin{tabular}{|l|lll|l|lll|}
\toprule
                                                                                            Expression & Support $x_1$ & Support $x_2$ & Support $x_3$ &                                                          Expression & Support $x_1$ & Support $x_2$ & Support $x_3$ \\
\midrule
                                             $\frac{\sqrt{2} e^{- \frac{x_{1}^{2}}{2}}}{2 \sqrt{\pi}}$ &    (1, 3) &      None &      None &  $\frac{x_{1} x_{3}^{2}}{\sqrt{- \frac{x_{2}^{2}}{x_{3}^{2}} + 1}}$ &    (1, 5) &    (1, 2) &   (3, 10) \\
                             $\frac{\sqrt{2} e^{- \frac{x_{2}^{2}}{2 x_{1}^{2}}}}{2 \sqrt{\pi} x_{1}}$ &    (1, 3) &    (1, 3) &      None &                                     $\frac{x_{1}}{4 \pi x_{2}^{2}}$ &    (1, 5) &    (1, 5) &      None \\
        $\frac{\sqrt{2} e^{- \frac{\left(x_{2} - x_{3}\right)^{2}}{2 x_{1}^{2}}}}{2 \sqrt{\pi} x_{1}}$ &    (1, 3) &    (1, 3) &    (1, 3) &                                   $\frac{x_{1}}{4 \pi x_{2} x_{3}}$ &    (1, 5) &    (1, 5) &    (1, 5) \\
                                              $\frac{x_{1}}{\sqrt{- \frac{x_{2}^{2}}{x_{3}^{2}} + 1}}$ &    (1, 5) &    (1, 2) &   (3, 10) &                            $\frac{3 x_{1}^{2}}{20 \pi x_{2} x_{3}}$ &    (1, 5) &    (1, 5) &    (1, 5) \\
                                                                                         $x_{1} x_{2}$ &    (1, 5) &    (1, 5) &      None &                                         $\frac{x_{1} x_{2}^{2}}{2}$ &    (1, 5) &    (1, 5) &      None \\
                                                                 $\frac{x_{1}}{4 \pi x_{2} x_{3}^{2}}$ &    (1, 5) &    (1, 5) &    (1, 5) &                        $\frac{x_{1}}{x_{2} \left(x_{3} + 1\right)}$ &    (1, 5) &    (1, 5) &    (1, 5) \\
                                                                                         $x_{1} x_{2}$ &    (1, 5) &    (1, 5) &      None &               $\frac{x_{1} x_{2}}{- \frac{x_{1} x_{2}}{3} + 1} + 1$ &    (0, 1) &    (0, 1) &      None \\
                                                                                   $x_{1} x_{2} x_{3}$ &    (1, 5) &    (1, 5) &    (1, 5) &            $\frac{x_{1}}{\sqrt{- \frac{x_{2}^{2}}{x_{3}^{2}} + 1}}$ &    (1, 5) &   (1, 2) &   (3, 10) \\
                                                                           $\frac{x_{1}^{2} x_{2}}{2}$ &    (1, 5) &    (1, 5) &      None &      $\frac{x_{1} x_{2}}{\sqrt{- \frac{x_{2}^{2}}{x_{3}^{2}} + 1}}$ &    (1, 5) &    (1, 2) &   (3, 10) \\
                                        $\frac{x_{1} x_{2}}{\sqrt{- \frac{x_{2}^{2}}{x_{3}^{2}} + 1}}$ &    (1, 5) &    (1, 2) &   (3, 10) &                           $- x_{1} x_{2} \cos{\left(x_{3} \right)}$ &    (1, 5) &    (1, 5) &    (1, 5) \\
                                             $\frac{x_{2} + x_{3}}{1 + \frac{x_{2} x_{3}}{x_{1}^{2}}}$ &    (1, 5) &    (1, 5) &    (1, 5) &                           $- x_{1} x_{2} \cos{\left(x_{3} \right)}$ &    (1, 5) &    (1, 5) &    (1, 5) \\
                                                               $x_{1} x_{2} \sin{\left(x_{3} \right)}$ &    (1, 5) &    (1, 5) &    (0, 5) &    $\sqrt{\frac{x_{1}^{2}}{x_{2}^{2}} - \frac{\pi^{2}}{x_{3}^{2}}}$ &    (4, 6) &    (1, 2) &    (2, 4) \\
                                                                                 $\frac{x_{1}}{x_{2}}$ &    (1, 5) &    (1, 5) &      None &                                             $x_{1} x_{2} x_{3}^{2}$ &    (1, 5) &    (1, 5) &    (1, 5) \\
                                  $\operatorname{asin}{\left(x_{1} \sin{\left(x_{2} \right)} \right)}$ &    (0, 1) &    (1, 5) &      None &                                                   $x_{1} x_{2}^{2}$ &    (1, 5) &    (1, 5) &      None \\
                                                     $\frac{1}{\frac{x_{3}}{x_{2}} + \frac{1}{x_{1}}}$ &    (1, 5) &    (1, 5) &    (1, 5) &                                   $\frac{x_{1} x_{2}}{2 \pi x_{3}}$ &    (1, 5) &    (1, 5) &    (1, 5) \\
                                                                                 $\frac{x_{1}}{x_{2}}$ &   (1, 10) &   (1, 10) &      None &                                       $\frac{x_{1} x_{2} x_{3}}{2}$ &    (1, 5) &    (1, 5) &    (1, 5) \\
 $\frac{x_{1} \sin^{2}{\left(\frac{x_{2} x_{3}}{2} \right)}}{\sin^{2}{\left(\frac{x_{2}}{2} \right)}}$ &    (1, 5) &    (1, 5) &    (1, 5) &                                   $\frac{x_{1} x_{2}}{4 \pi x_{3}}$ &    (1, 5) &    (1, 5) &    (1, 5) \\
                                        $\operatorname{asin}{\left(\frac{x_{1}}{x_{2} x_{3}} \right)}$ &    (1, 2) &    (2, 5) &    (1, 5) &                                $x_{1} x_{2} \left(x_{3} + 1\right)$ &    (1, 5) &    (1, 5) &    (1, 5) \\
                                                               $\frac{x_{3}}{1 - \frac{x_{2}}{x_{1}}}$ &   (3, 10) &    (1, 2) &    (1, 5) &                                         $\frac{x_{1}}{2 x_{2} + 2}$ &    (1, 5) &    (1, 5) &      None \\
           $\frac{x_{3} \left(1 + \frac{x_{2}}{x_{1}}\right)}{\sqrt{1 - \frac{x_{2}^{2}}{x_{1}^{2}}}}$ &   (3, 10) &    (1, 2) &    (1, 5) &                                   $\frac{4 \pi x_{1} x_{2}}{x_{3}}$ &    (1, 5) &    (1, 5) &    (1, 5) \\
                                                                           $\frac{x_{1} x_{2}}{2 \pi}$ &    (1, 5) &    (1, 5) &      None &           $\sin^{2}{\left(\frac{2 \pi x_{1} x_{2}}{x_{3}} \right)}$ &    (1, 2) &    (1, 2) &    (1, 4) \\
                                      $x_{1} + x_{2} + 2 \sqrt{x_{1} x_{2}} \cos{\left(x_{3} \right)}$ &    (1, 5) &    (1, 5) &    (1, 5) &                                         $\frac{x_{1} x_{2}}{2 \pi}$ &    (1, 5) &    (1, 5) &      None \\
                                                                             $\frac{3 x_{1} x_{2}}{2}$ &    (1, 5) &    (1, 5) &      None &          $2 x_{1} \left(1 - \cos{\left(x_{2} x_{3} \right)}\right)$ &    (1, 5) &    (1, 5) &    (1, 5) \\
                                                                       $\frac{x_{2} x_{3}}{x_{1} - 1}$ &    (2, 5) &    (1, 5) &    (1, 5) &                       $\frac{x_{1}^{2}}{8 \pi^{2} x_{2} x_{3}^{2}}$ &    (1, 5) &    (1, 5) &    (1, 5) \\
                                                                                   $x_{1} x_{2} x_{3}$ &    (1, 5) &    (1, 5) &    (1, 5) &                                   $\frac{2 \pi x_{1}}{x_{2} x_{3}}$ &    (1, 5) &    (1, 5) &    (1, 5) \\
                                                                    $\sqrt{\frac{x_{1} x_{2}}{x_{3}}}$ &    (1, 5) &    (1, 5) &    (1, 5) &            $x_{1} \left(x_{2} \cos{\left(x_{3} \right)} + 1\right)$ &    (1, 5) &    (1, 5) &    (1, 5) \\
\bottomrule
\end{tabular}
}
\endgroup

\caption{AI-Feynman equation with less than 4 independent variables and the supports as indicated in \protect{\cite{AIf1}}}
\label{tab:ai_fey}
\end{table}%

\begin{table}[h]
\centering
\begingroup
\setlength{\tabcolsep}{10pt} %
\renewcommand{\arraystretch}{2.} %
\begin{tabular}{|l|l|}
\toprule
                                                                                          Expression &                                                                                            Expression \\
\midrule
                           $4.931 x_{1} - x_{2} + 4.023 \tan{\left(x_{1}^{2} - 4.027 x_{3} \right)}$ &                                                    $x_{1} \sqrt{- x_{3} + \sin{\left(x_{2} \right)}}$ \\
 $\sin{\left(\cos{\left(3.488 x_{1} \tan{\left(2.798 x_{1} \right)} + 2.938 x_{1} \right)} \right)}$ &                $2.29 x_{2} \cos{\left(x_{2} \right)} + \cos{\left(\frac{1.044 x_{1}}{x_{2}} \right)}$ \\
                                                        $\sqrt{- x_{1} + \frac{x_{2} x_{3}}{x_{1}}}$ &                                 $\frac{x_{3} + \frac{3.797 \sin{\left(x_{1} \right)}}{x_{2}}}{x_{3}}$ \\
                           $\sin{\left(4.84 x_{3} \left(2.3 x_{1} - 3.494 x_{2} + 1\right) \right)}$ &                                       $x_{1} - 4.843 x_{2} x_{3} + x_{2} + \cos{\left(x_{3} \right)}$ \\
                       $\sin{\left(x_{3} \right)} + \sin{\left(\frac{x_{3}}{x_{1} - x_{2}} \right)}$ &                            $\cos{\left(x_{1} + 1.504 x_{2} + \left(x_{2} + x_{3}\right)^{2} \right)}$ \\
                              $x_{1} \left(2.683 x_{1} + x_{2} \cos{\left(x_{3} \right)}\right) + 1$ &                       $x_{1} \left(- 4.641 x_{1} + \cos^{2}{\left(4.959 \sqrt{x_{2}} \right)}\right)$ \\
                 $4.631 \sin{\left(4.419 \sin{\left(\frac{x_{2} x_{3}}{x_{1}^{2}} \right)} \right)}$ &                                                $x_{2} \left(x_{2} - \frac{- x_{1} - 1}{x_{2}}\right)$ \\
                                          $3.874 x_{3} + 4.12 - \frac{1}{x_{1} + 4.322 x_{2} x_{3}}$ &                                          $4.47 x_{1} + 1.193 \cos{\left(1 + \frac{1}{x_{2}} \right)}$ \\
                                           $\frac{1.858 x_{1} x_{3}}{- x_{1} + x_{2}} - 3.661 x_{3}$ &                                              $3.63 x_{1} \cos{\left(1.427 x_{2}^{3} + x_{2} \right)}$ \\
                                         $2.846 x_{2} + \sin{\left(x_{1}^{5} + 2.258 x_{3} \right)}$ &                                $- x_{1} \left(1.196 x_{1} + \sin{\left(x_{1} + x_{2} \right)}\right)$ \\
                                         $\frac{x_{2} + x_{3} + \frac{x_{2} - 4.615}{x_{2}}}{x_{1}}$ &                                                   $x_{3} + \frac{x_{3}}{x_{1} + 2.318 x_{2} + x_{3}}$ \\
                $- x_{3} + \frac{0.221 \left(- x_{1} + x_{2}\right)}{\log{\left(x_{2} \right)}} - 1$ &                                               $x_{1} - \frac{7.74 \sqrt{0.383 x_{1} + x_{2}}}{x_{2}}$ \\
             $\left(1.261 x_{1} + 3.29 \cos{\left(1 \right)}\right) \log{\left(4.169 x_{2} \right)}$ &          $1 + \frac{0.221 \tan{\left(3.972 x_{2} \right)}}{x_{2} \left(- 3.549 x_{1} + x_{2}\right)}$ \\
                                             $\frac{x_{2}}{x_{2} + \cos{\left(x_{1} x_{3} \right)}}$ &                        $1 - \sin{\left(x_{1} \left(x_{1} + \sin{\left(x_{1} \right)}\right) \right)}$ \\
                                             $2.161 x_{3} \cos^{3}{\left(x_{1}^{2} + x_{2} \right)}$ &                                        $\cos{\left(x_{1} \right)} - \sqrt{\cos{\left(x_{2} \right)}}$ \\
             $3.196 \tan{\left(\cos{\left(4.459 x_{1} \right)} - \tan{\left(1 \right)} \right)} - 1$ &                                   $- 2.586 x_{2} + \frac{0.693 \cos{\left(x_{2} - 1 \right)}}{x_{1}}$ \\
                                                        $\sqrt{x_{2}^{2} - x_{2} - e^{2.103 x_{1}}}$ &                                         $- 8.802 x_{1} + 3.379 \log{\left(x_{1} + x_{2}^{4} \right)}$ \\
                        $- x_{2} + \log{\left(x_{1} \left(- x_{2} + \frac{1.513}{e}\right) \right)}$ &                               $\sin{\left(\frac{2.696 x_{2}}{- x_{1} + 2.364 x_{2}} \right)} + 1.097$ \\
                      $3.919 \log{\left(1.731 \sqrt{x_{1} \sin{\left(2.176 x_{2} \right)}} \right)}$ &                                                         $x_{1} - x_{2} \cos{\left(x_{1}^{3} \right)}$ \\
             $x_{1} + \frac{0.422 x_{1}}{x_{2} \left(4.26 x_{1} + \cos{\left(x_{2} \right)}\right)}$ &                    $x_{1} - 2.636 \left(3.271 x_{2} + x_{3}\right) \sin{\left(x_{1} \right)} + 2.387$ \\
      $\sin{\left(\frac{3.553 \left(- 0.531 x_{2} + x_{3}\right)^{2}}{x_{1} + 1.244 x_{2}} \right)}$ &                                                 $x_{1} \left(x_{1} + x_{3} + 1\right) + \sqrt{x_{2}}$ \\
                                             $\sqrt{x_{1} \left(\frac{x_{2}}{x_{3}} + x_{3}\right)}$ &              $x_{2} + \log{\left(3.949 x_{1} \left(x_{1} + \sin{\left(x_{2} \right)}\right) \right)}$ \\
                               $- x_{1} \sin{\left(\tan{\left(x_{1} x_{2} \right)} \right)} + x_{1}$ &            $x_{1} + 3.183 \sin{\left(3.696 \log{\left(x_{2} \left(x_{2} - 1\right) \right)} \right)}$ \\
                                                  $x_{2} \sin{\left(x_{1} - 1.47 x_{3}^{2} \right)}$ &                                                   $x_{2}^{2} + 1.209 \sin{\left(x_{1} x_{2} \right)}$ \\
                           $x_{1} \log{\left(\sin{\left(\tan{\left(x_{1} \right)} \right)} \right)}$ &  $\frac{x_{2} \left(x_{1} + 1.756 \log{\left(2.756 \cos{\left(x_{3} \right)} \right)}\right)}{x_{1}}$ \\
\bottomrule
\end{tabular}

\endgroup

\caption{50 random equations extracted from the SOOBE dataset  (version with constants). }
\label{tab:SOOBE}
\end{table}%

\FloatBarrier

\section{Additional Results}\label{app:res}

\subsection{Additional Metrics on all Benchmarks}
In this section, we show that the conclusions drawn in Section \ref{sec:results} with the $A_2$ metric are consistent when the $A_1$ metric is considered instead.

\begin{figure*}[h]
    \centering
    \includegraphics[width=0.93\linewidth]{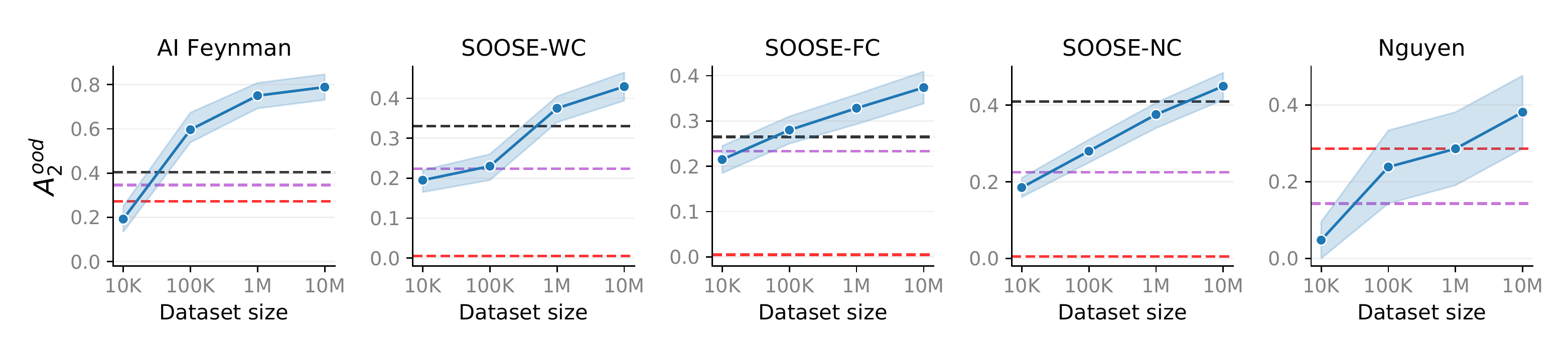}
    \includegraphics[width=0.6\linewidth]{figs/plots_2021_june_camera_ready/training_amounts__legend.pdf}
    \caption{Accuracy as a function of the size of the pre-training dataset, for a fixed computational budget ($\sim$100 s) at test time.} %
        
    \label{fig:training_amounts}
\end{figure*}

\begin{figure*}[h]
    \centering
     \includegraphics[width=0.93\linewidth]{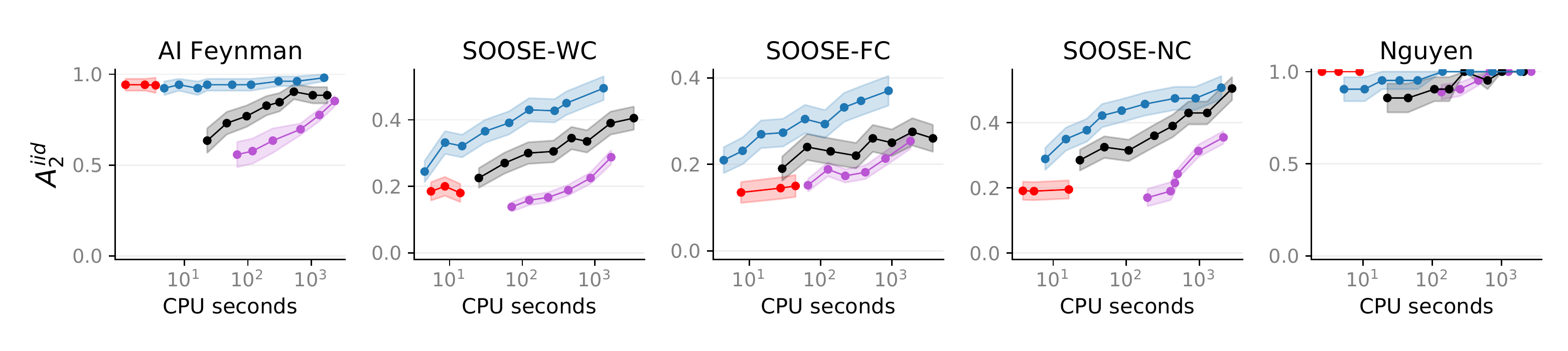}%
    \\
    \includegraphics[width=0.65\linewidth]{figs/plots_2021_june_camera_ready/acc_vs_time/acc_vs_time_5cols_legend.pdf}
    \caption{Accuracy in distribution as a function of time for all methods ran on a single CPU per equation.}
    \label{fig:acc_vs_time_ours_iid2}
\end{figure*}

\begin{figure*}[h]
    \centering
     \includegraphics[width=0.93\linewidth]{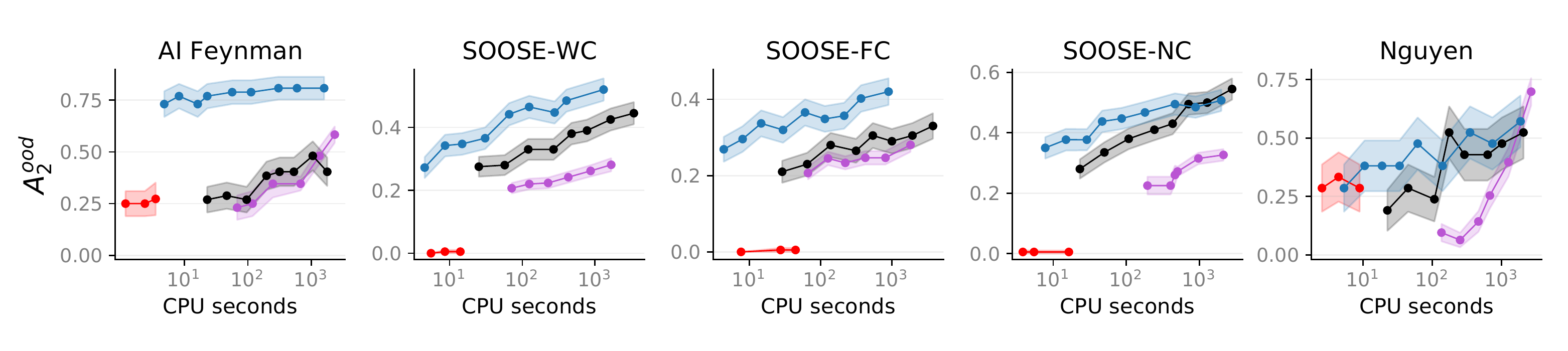}%
    \\
    \includegraphics[width=0.65\linewidth]{figs/plots_2021_june_camera_ready/acc_vs_time/acc_vs_time_5cols_legend.pdf}
    \caption{Accuracy out of distribution as a function of time for all methods ran on a single CPU per equation.}
    \label{fig:acc_vs_time_ours_ood2}
\end{figure*}

\begin{figure}[H]
    \centering
    \includegraphics[width=0.2\linewidth]{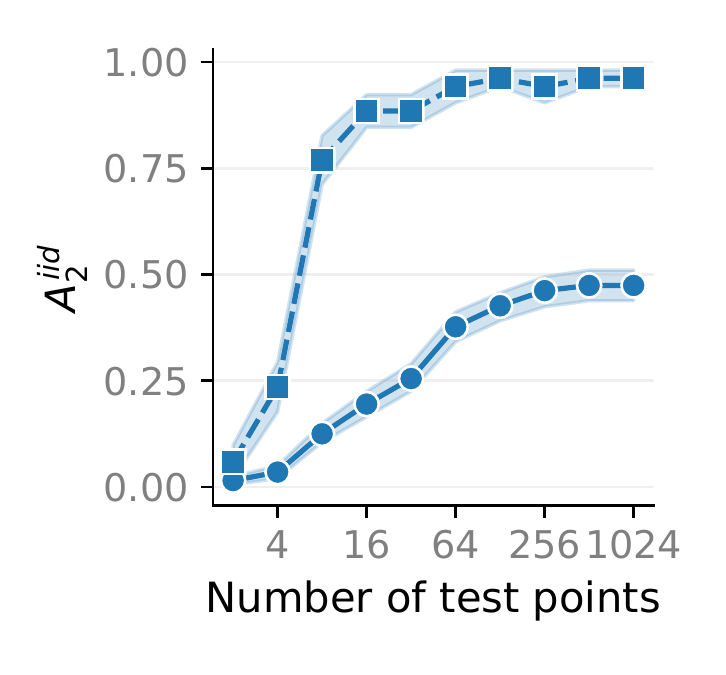}
    \includegraphics[width=0.2\linewidth]{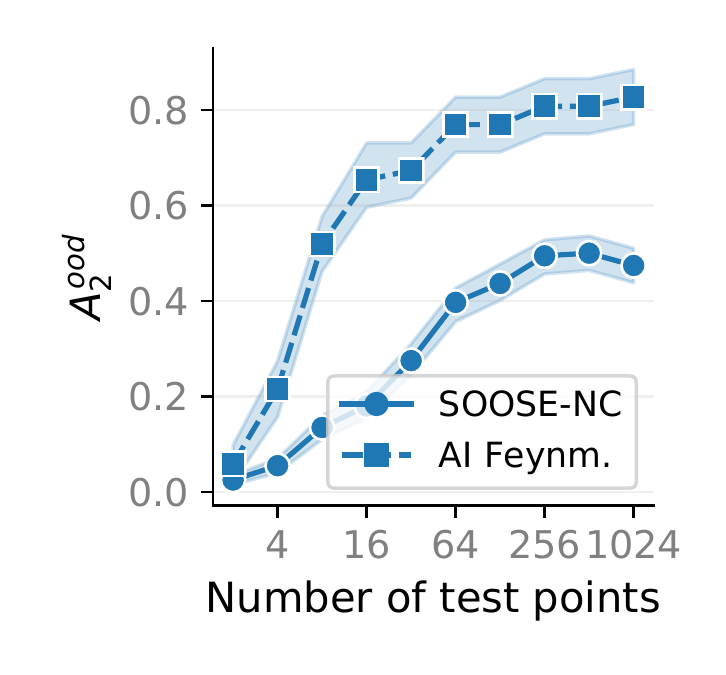}%
        \caption{Accuracy as a function of number of input-output pairs observed at test time.%
        }
    \label{fig:acc_vs_points_A2}
\end{figure}

\subsection{OOD Extrapolation Examples}
\begin{figure*}[h]
\centering
     \includegraphics[align=t,width=0.99\linewidth]{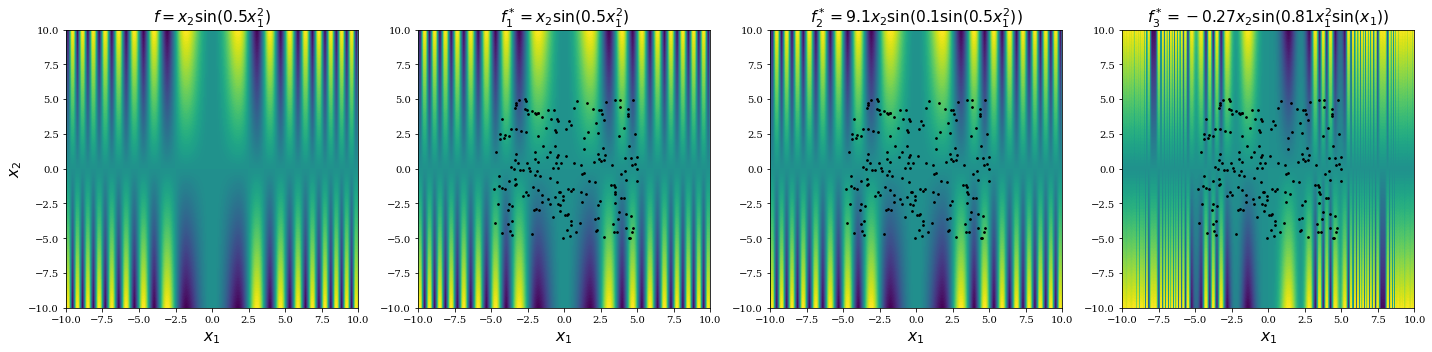}%
     \includegraphics[align=t,width=0.046\linewidth]{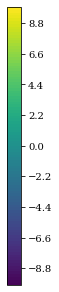}\\
     \includegraphics[align=t,width=0.99\linewidth]{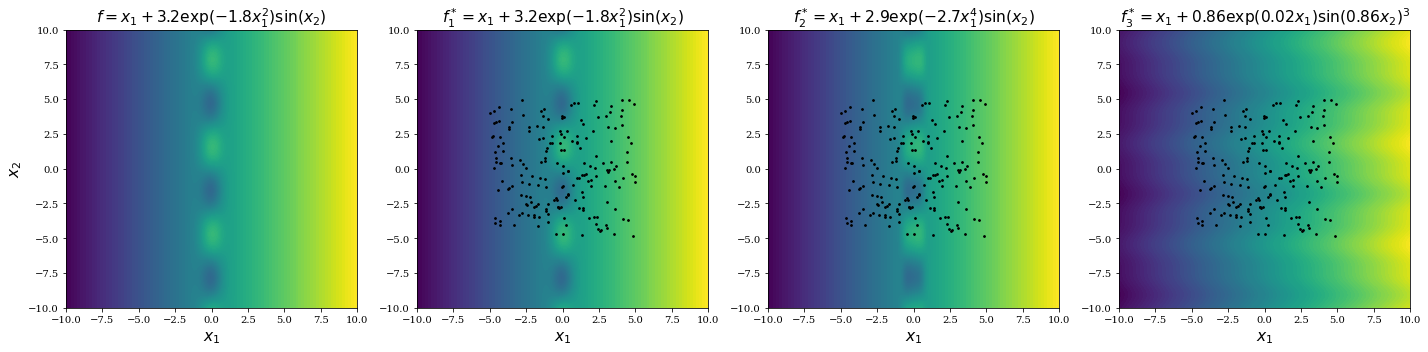}%
     \includegraphics[align=t,width=0.046\linewidth]{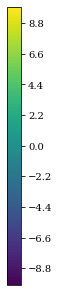}\\
     \includegraphics[align=t,width=0.99\linewidth]{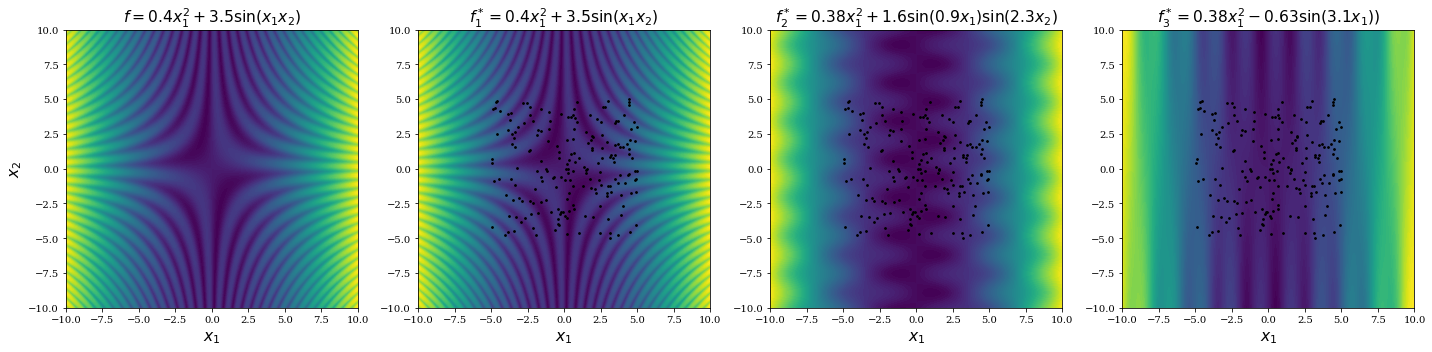}%
     \includegraphics[align=t,width=0.040\linewidth]{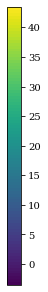}\\
     \includegraphics[align=t,width=0.99\linewidth]{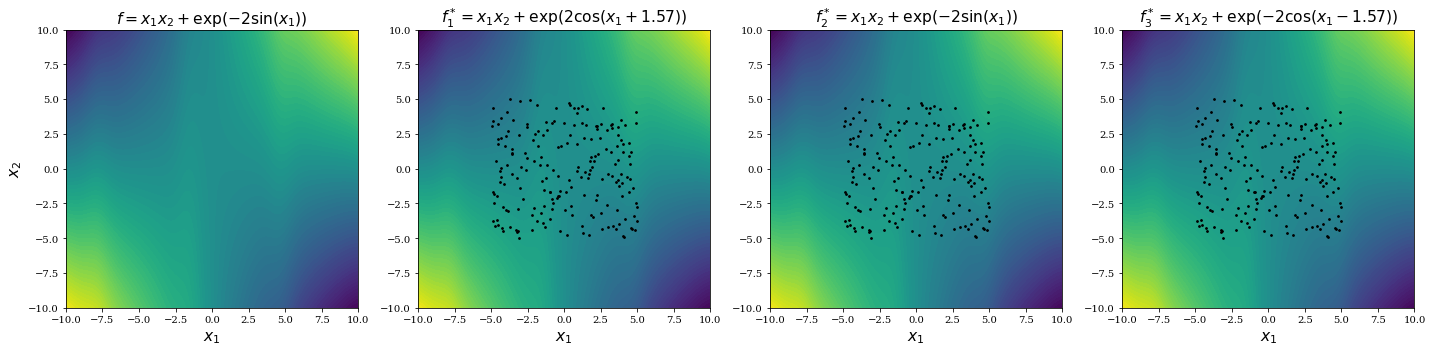}%
     \includegraphics[align=t,width=0.051\linewidth]{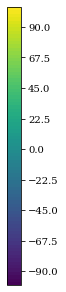}\\
    \caption{Four examples of functions learned by our model. The first column shows the ground truth equation, whereas the remaining three represent the top predictions of our model sorted by likelihood (from left to right).}\label{fig::2d}
\end{figure*}

\paragraph{Examples of 2D Functions}
Fig. \ref{fig::2d} provides 4 examples of functions learned by our model. All the considered functions depend only on two independent variables, $x_1$ and $x_2$. The visualizations show that our method, given a relatively small set of support points (black dots in the figure) is able to extrapolate out of distribution by retrieving the underlying symbolic expression. 
The last row of Fig. \ref{fig::2d} shows that NeSymReS at times finds valid alternative expressions based on trigonometric identities, i.e. $\sin(x) = \cos(x-\pi/2) = -\cos(x+\pi/2)$.

\end{document}